\documentclass[conference]{IEEEtran}
\usepackage{amsmath,amsfonts}
\usepackage{algorithmic}
\usepackage{algorithm}
\usepackage{array}
\usepackage{multirow}
\usepackage{booktabs}
\usepackage{tikz}
\usetikzlibrary{positioning,decorations.pathreplacing,calc}
\usepackage{colortbl} 
\usepackage{pifont}   
\usepackage[caption=false,font=normalsize,labelfont=sf,textfont=sf]{subfig}
\usepackage{textcomp}
\usepackage{stfloats}
\usepackage{url}
\usepackage{verbatim}
\usepackage{graphicx}
\usepackage{cite}
\usepackage[breaklinks=true,hidelinks]{hyperref}
\usepackage{cuted}
\usepackage{capt-of} 

\newcommand{\vc}[1]{\boldsymbol{#1}}          

\newcommand{\Rmat}{\vc{R}}
\newcommand{\norm}[1]{\lVert #1 \rVert}
\newcommand{\sqnorm}[1]{\lVert #1 \rVert^2}
\newcommand{\sqfro}[1]{\lVert #1 \rVert_F^2}   
\definecolor{markgreen}{RGB}{0,140,60}
\definecolor{markred}{RGB}{200,30,30}
\newcommand{\cmark}{\textcolor{markgreen}{\ding{51}}}   
\newcommand{\xmark}{\textcolor{markred}{\ding{55}}}   

\begin{document}
\bstctlcite{IEEEexample:BSTcontrol}

\title{HOI-Retarget: Contact-Centric Retargeting\\ for Human-Object Interaction}

\author{
\IEEEauthorblockN{Jihwan Shin, Adri\`a L\'opez Escoriza, Junzhe He, Matthias Heyrman, Marco Hutter}
\IEEEauthorblockA{Robotic Systems Lab, ETH Z\"urich, 8092 Z\"urich, Switzerland\\
\{jishin, alopez, junzhe, mheyrman, mahutter\}@ethz.ch}
}

\maketitle

\begin{strip}
    \centering
    \vspace{-4.5em}
    \includegraphics[width=\textwidth]{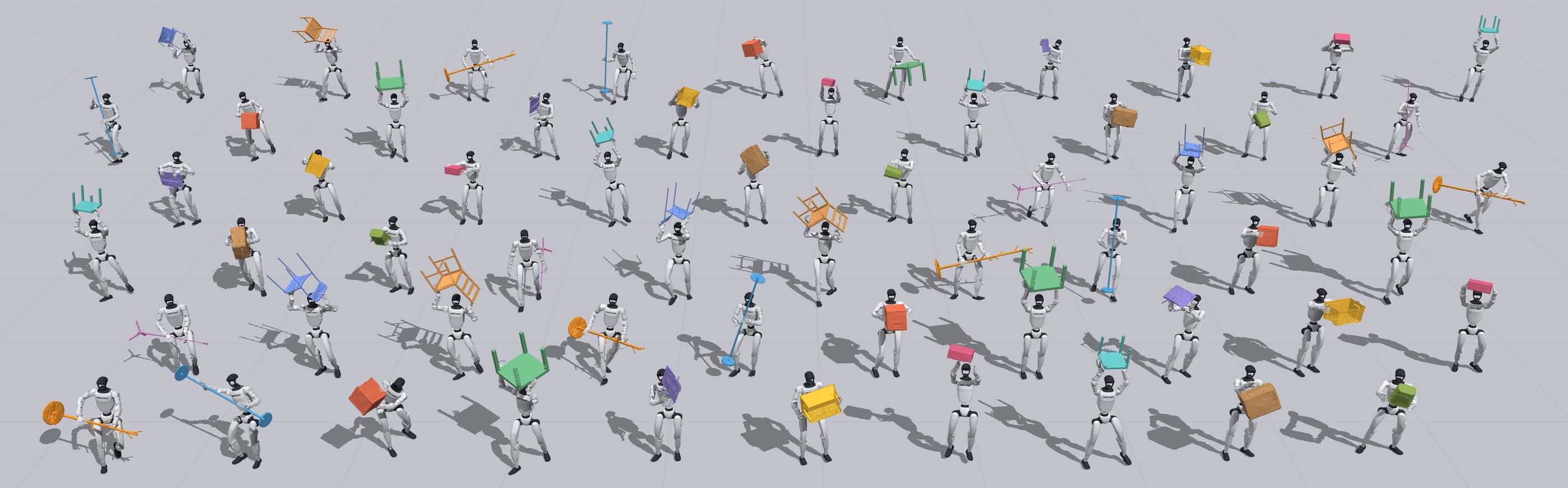}
    \captionof{figure}{HOI-Retarget preserves object-relative contacts while generating robot interaction references across diverse motions, objects and humanoid platforms.}
    \label{fig:title}
\end{strip}

\begin{abstract}

Learning from demonstration (LfD) has enabled humanoid robots to acquire diverse whole-body skills, but extending this paradigm to human-object interaction (HOI) is limited by the availability of robot-compatible interaction references.
We present HOI-Retarget, a contact-centric retargeting method that transfers HOI onto a humanoid robot for large-scale motion-data generation.
Its windowed trajectory optimization uses every labeled contact as a target in the object frame, balancing body tracking, foot support and smoothness under the robot's kinematic limits. 
The method can augment a single demonstration across object sizes, absorb contacts reconstructed from monocular video, and extend to several robots manipulating one object.
We publicly release the code and the retargeted motion dataset. 
Website: \mbox{\textbf{\href{https://shinben0327.github.io/hoi-retarget}{\textcolor{red}{shinben0327.github.io/hoi-retarget}}}.}
\end{abstract}

\section{Introduction}
\label{sec:introduction}

\begin{figure*}[!t]
        \centering
        \includegraphics[width=\linewidth]{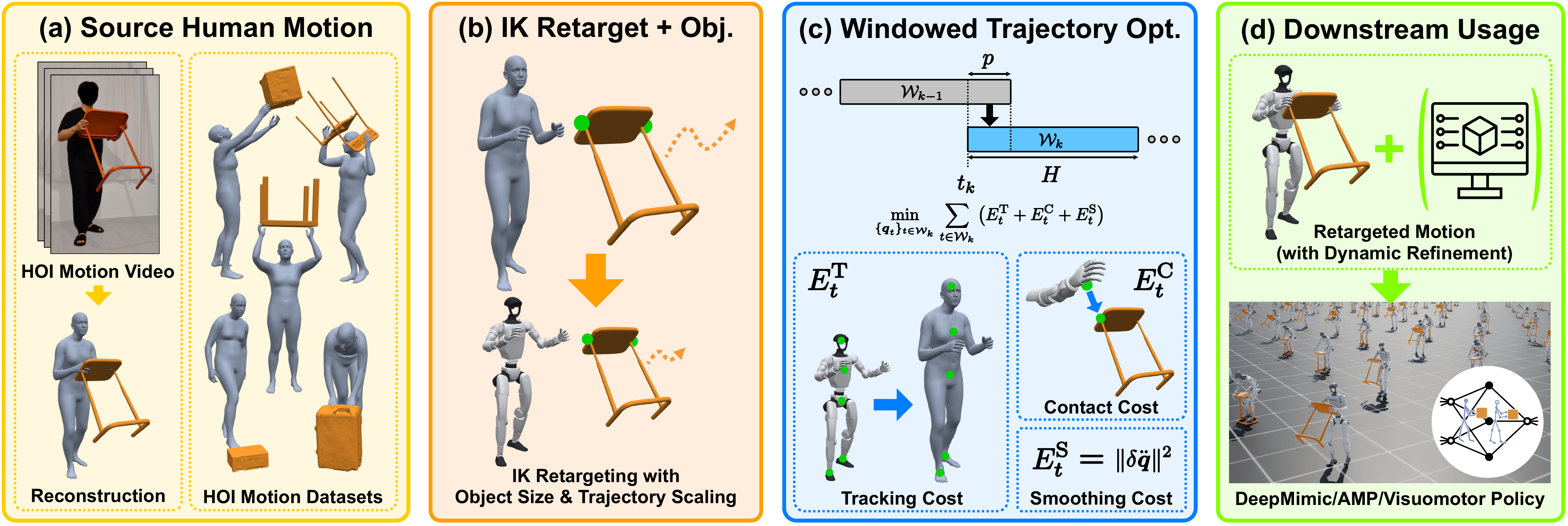}
        \caption{\textbf{Overview of HOI-Retarget}. (a) The source is a captured or video-reconstructed HOI clip, providing human motion, an object trajectory and contact labels. (b) IK retargeting maps the human onto the robot, while the object mesh and trajectory are scaled by the robot-to-human height ratio, carrying the contact targets with them. (c) A windowed trajectory optimization with tracking, contact and smoothness costs recovers those contacts under the robot's kinematic limits. (d) The result drives downstream policies directly, or after dynamic refinement in simulation.}
        \label{fig:summary}
\end{figure*}

Humanoid robots are well suited to environments designed for people, but useful deployment requires more than locomotion: robots must coordinate their whole body while making and breaking contact with objects.
Learning from demonstration (LfD) offers a practical route to such loco-manipulation skills by using demonstrated motion to guide exploration~\cite{peng2018deepmimic, peng2021amp, xu2025intermimic}.
The breadth of the resulting controller, however, depends on the availability of diverse, robot-compatible interaction references.
Collecting these references directly on hardware through teleoperation~\cite{he2024h2o, luo2026sonic, ze2025twist2} is labor-intensive, whereas motion-capture datasets and video provide a growing source of human-object interaction (HOI) data~\cite{li2023omomo,lu2025humoto,cari4d2026}.
Turning this human data into usable robot motion is therefore an important step toward scaling humanoid loco-manipulation.

Retargeting HOI is not merely a body-pose transfer problem.
Because humans and humanoids differ in proportions, joint structure, and reachable workspace, independently matching their poses can move a hand away from the object or shift it to a functionally different surface.
These errors alter the interaction itself and produce poor references for downstream control.
General humanoid retargeting methods primarily align body poses and end effectors~\cite{luo2023phc,araujo2025gmr,kim2025pyroki}, while interaction-aware methods such as OmniRetarget~\cite{yang2025omniretarget} preserve broader body--object--environment relationships through interaction-mesh deformation.
However, preserving the overall interaction geometry does not directly minimize the error to each source contact location, which becomes critical for objects with thin or spatially localized grasp regions.
Physics-based methods can subsequently repair dynamic infeasibility~\cite{dhedin2026dynaretarget,pan2025spider}, but still benefit from kinematic references that already encode the intended object interaction accurately.

We present \textbf{HOI-Retarget}, a contact-centric method that transfers human HOI to humanoid embodiments while explicitly preserving the location and timing of object contacts (Fig.~\ref{fig:summary}).
Starting from an inverse-kinematics (IK) retarget, we scale the object and its trajectory to the robot embodiment and express each active source contact as a target in the object frame.
A trajectory optimization over overlapping windows then balances body-motion tracking, contact alignment, foot support, and temporal smoothness under joint-position and velocity limits.
This formulation couples consecutive frames while keeping computation tractable for long clips, and naturally preserves contact targets under object-size augmentation.
For contact annotations corrupted by monocular reconstruction, an optional interactive tool allows the target location and interval of each contact segment to be corrected before optimization.

Across OMOMO~\cite{li2023omomo} clips with 13 object categories retargeted to a Unitree G1~\cite{unitree_g1}, HOI-Retarget reduces the mean source contact-point gap from $18.3$~cm for the current state-of-the-art interaction-aware method~\cite{yang2025omniretarget} to $0.5$~cm, while running over $4\times$ faster.
We open-source the pipeline and a dataset of 6,952 robot HOI clips drawn from five source datasets \cite{li2023omomo, kim2024parahome, zhang2023neuraldome, zuo2026corolehoi, zhao2024imhd}, totaling 13.8 hours across 75 unique objects, with retargets for multiple humanoid platforms \cite{unitree_g1, unitree_h2} and synchronized multi-robot collaboration motions.

Our contributions are as follows:
\begin{enumerate}
    \item A contact-centric HOI retargeting formulation that directly preserves object-frame contact targets through temporally coupled windowed trajectory optimization.
    \item A scalable data-generation pipeline that supports different humanoid embodiments and object scales, together with optional contact correction for noisy reconstructed interactions.
    \item An open-source dataset of 6,952 robot HOI clips spanning 75 unique objects, five source datasets, humanoid embodiments, and single- and multi-robot interactions.
\end{enumerate}

\section{Related Work}
\label{sec:related_work}

\subsection{Human-to-Robot Motion Retargeting}

Humanoid motion retargeting commonly optimizes robot configurations to match human body positions or orientations while respecting embodiment-specific limits.
PHC~\cite{luo2023phc} uses keypoint-based optimization to prepare human motion for physics-based character control, while GMR~\cite{araujo2025gmr} provides a fast inverse-kinematics pipeline and demonstrates that reference quality strongly affects downstream tracking.
PyRoki~\cite{kim2025pyroki} offers a modular formulation of kinematic objectives and constraints for robot retargeting.
Recent methods also learn mappings from human to robot motion: NMR~\cite{zhao2026nmr} learns to project motion onto a dynamics-aware robot distribution, while ReActor~\cite{muller2026reactor} couples retargeting with reinforcement learning (RL) based physical execution.
These approaches improve body-motion fidelity or feasibility, but do not explicitly represent the location and duration of contacts with a manipulated object.

\subsection{Interaction-Aware Retargeting}

OmniRetarget~\cite{yang2025omniretarget} is the closest kinematic method to ours.
It deforms an interaction mesh constructed from body, object, and environment points while enforcing kinematic constraints, thereby preserving broad spatial relationships and enabling augmentation across terrains, objects, and robot embodiments.
HOI-Retarget (ours) instead uses labeled object contacts as its primary interaction representation and directly minimizes their object-frame error over temporal windows.
This narrower formulation targets accurate object manipulation rather than general scene interaction and permits individual contact segments to be inspected and corrected.
Table~\ref{tab:related_comparison} summarizes how these methods differ from HOI-Retarget.

DynaRetarget~\cite{dhedin2026dynaretarget} uses sampling-based trajectory optimization (SBTO) to transform imperfect robot--object references into dynamically feasible motions, SPIDER~\cite{pan2025spider} uses large-scale physics sampling and virtual contact guidance to refine and augment kinematic demonstrations across humanoid and dexterous embodiments, and ULTRA~\cite{he2026ultra} trains a physics-driven HOI retargeting policy at dataset scale.
These methods address the complementary dynamic-retargeting stage: they optimize physical execution from an existing reference, whereas our focus is generating a contact-accurate kinematic reference from human HOI.
The output of HOI-Retarget can consequently provide an interaction-preserving initialization for such physics-based refinement.

\subsection{Human-Object Interaction Data}

Captured and curated HOI datasets provide synchronized human and object trajectories across increasingly diverse interactions, including OMOMO~\cite{li2023omomo}, CORE4D~\cite{liu2025core4d}, HUMOTO~\cite{lu2025humoto}, and InterAct~\cite{xu2025interact}.
Our primary experiments use OMOMO trajectories packaged by InterMimic~\cite{xu2025intermimic} and per-body contact annotations originating from InterAct.
Video-based methods can broaden this source beyond dedicated capture systems: NeuralDome~\cite{zhang2023neuraldome} and I'm-HOI~\cite{zhao2024imhd} reconstruct coupled human--object motion, while CARI4D~\cite{cari4d2026} recovers category-agnostic HOI from monocular RGB video and reasons explicitly about contact.
Nevertheless, captured and reconstructed motions remain human-specific, and estimated contacts can be displaced by occlusion, depth ambiguity, or body--object penetration.
HOI-Retarget converts these heterogeneous sources into a common robot representation and exposes object-frame contact segments that can be refined when the source estimate is unreliable.

\begin{table*}[!t]
\centering
\caption{Capabilities of representative retargeting methods.}
\label{tab:related_comparison}
\scriptsize
\setlength{\tabcolsep}{6pt}
\renewcommand{\arraystretch}{1.15}
\begin{tabular}{lcccccccc}
\toprule
Method & \shortstack{Source Human\\Motion} & \shortstack{Object\\Interaction} & \shortstack{Contact  Location\\Preservation} & \shortstack{Data\\Augmentation} & \shortstack{Multi-Agent\\Source} & \shortstack{Temporal\\Coupling} & \shortstack{Dynamics\\Enforcement} & \shortstack{Optimization\\Method} \\
\midrule
GMR~\cite{araujo2025gmr}                   & \cmark & \xmark & \xmark & \xmark & \xmark & \xmark & \xmark & per-frame IK \\
OmniRetarget~\cite{yang2025omniretarget}   & \cmark & \cmark & \xmark & \cmark & \xmark & \xmark & \xmark & per-frame SOCP \\
DynaRetarget~\cite{dhedin2026dynaretarget} & \xmark & \cmark & \xmark & \cmark & \xmark & \cmark & \cmark & progressive SBTO \\
SPIDER~\cite{pan2025spider}                & \xmark & \cmark & \xmark & \cmark & \xmark & \cmark & \cmark & receding-horizon sampling \\
\midrule
\textbf{HOI-Retarget (ours)}               & \cmark & \cmark & \cmark & \cmark & \cmark & \cmark & \xmark & windowed NLP \\
\bottomrule
\end{tabular}
\end{table*}
\section{HOI-Retarget}
\label{sec:hoi_retarget}

HOI-Retarget converts a human HOI clip into a robot-object interaction trajectory in two stages.
The first stage (Sec.~\ref{subsec:ikretargetobjscale}) closes the embodiment gap kinematically: IK maps the human pose onto the robot, and the object is rescaled together with its trajectory so that the interaction stays within the reach of the robot.
The second stage (Sec.~\ref{subsec:windowedto}) refines that reference with a windowed trajectory optimization, which recovers the source contacts on the object while keeping the motion faithful to the human, smooth, and inside the robot's kinematic limits.

\subsection{Inverse-Kinematics Retargeting with Object Scaling}
\label{subsec:ikretargetobjscale}

A source HOI clip provides a parametric SMPL-X human motion~\cite{smplx2019}, the object pose trajectory $(\vc p^w_o,\Rmat^w_o)$, and per-body human--object contact labels.
We bring every source into this common representation and label its contacts with a single geometric criterion~\cite{xu2025interact,xu2025intermimic}, so that datasets captured under different conventions are treated identically: a body is in contact when its mesh lies within a proximity threshold of the object, and a foot is in stance when its height and velocity are both under a certain threshold.
The labels are transferred onto the robot contact links $\mathcal C=\mathcal H\cup\mathcal F$, the two palms and the two feet, giving a contact flag $\beta_{i,t}$ for every $i\in\mathcal C$ and a stance flag $\gamma_{i,t}$ for every foot $i\in\mathcal F$.
At each contact frame we express the world position $\vc p^{w\mathrm h}_{c,i,t}$ of the corresponding human body in the object frame,
\begin{equation}
  \vc p^{o*}_{c,i,t} = \Rmat^{w\top}_{o,t}\big(\vc p^{w\mathrm h}_{c,i,t}-\vc p^{w}_{o,t}\big),
  \qquad \beta_{i,t}=1,
  \label{eq:contactpoint}
\end{equation}
which is the contact point the robot is later asked to reproduce.
Anchoring the target to the object rather than to the world makes it independent of where the object is carried, and, as used below, of how large the object is.

The human motion is retargeted into the robot configuration $\vc q=(\vc p^w_b,\Rmat^w_b,\vc\theta)$ through IK \cite{araujo2025gmr}, which fits the human skeleton to the robot by a root scale given by the robot-to-human height ratio and replants the lowest foot on the ground.
We apply the same ratio to the object, shrinking its geometry about its own origin and its trajectory about the world origin, so that the object stays within reach of the robot.
Because the contact targets \eqref{eq:contactpoint} are expressed in the object frame, they shrink with the geometry, and each one stays fixed on the object surface.
We exploit this for data augmentation: an additional multiplier on top of the default ratio rescales the object further, so a single source clip yields as many robot clips as desired, each preserving the interaction on a differently sized object (Sec.~\ref{subsec:dataaug}).
We then apply a Savitzky--Golay filter, which removes reconstruction noise as well as the set-down discontinuities that rescaling would otherwise introduce in the trajectory.
The object trajectory remains fixed during Sec.~\ref{subsec:windowedto}. 

\subsection{Windowed Trajectory Optimization}
\label{subsec:windowedto}

IK retargeting reproduces the human pose but not the interaction.
Shorter arms and a different joint layout displace the palms from the object, and solving each frame independently leaves jitter and floating feet behind.
We therefore refine the trajectory with a nonlinear program built on a differentiable rigid-body model~\cite{carpentier2019pinocchio} and solved by an interior-point method~\cite{andersson2019casadi,wachter2006ipopt}, warm-started from the IK solution.
The only decision variables are the configurations ${\vc q_t}$, with joint velocities following as finite differences of $\vc q$ rather than being optimized separately; the object trajectory enters as a parameter, so the interaction must be recovered entirely through the robot's own configuration.

\begin{subequations}
\label{eq:opti}
\begin{align}
  \min_{\{\vc q_t\}_{t\in\mathcal W_k}} \quad & \sum_{t \in \mathcal W_k}
    \big( E^{\mathrm T}_t + E^{\mathrm C}_t + E^{\mathrm S}_t \big)
    \label{eq:opti_obj}\\
  \mathrm{s.t.} \quad & \vc\theta_{\min}\le \vc\theta_t \le \vc\theta_{\max},
    \label{eq:opti_c1}\\[2pt]
  & \lvert \delta\vc\theta_t\rvert
      \le \dot{\vc\theta}_{\mathrm{lim}}\,\Delta t,
    \label{eq:opti_c2}\\[2pt]
  & \vc q_t = \vc q^{\,\mathcal W_{k-1}}_t, \quad t < t_k + p .
    \label{eq:opti_c3}
\end{align}
\end{subequations}

Instead of one problem spanning the whole clip, we solve the sequence of short overlapping windows $\mathcal W_k=[\,t_k,\,t_k+H\,)$ of $H$ frames given in \eqref{eq:opti}, subject to the joint position and velocity limits of the robot.
The first $p$ frames of each window are pinned to the trajectory already accumulated from the preceding windows \eqref{eq:opti_c3}, which keeps the concatenated motion continuous across window boundaries.
A window is long enough for the frames inside it to be coupled, so that jerk can be penalized and a contact resolved over the interval it spans, which a per-frame solve structurally cannot express. At the same time, it is short enough that the size of each program is independent of the length of the clip.
This choice is further examined in Appendix~\ref{app:horizon}.

Each of the three groups in \eqref{eq:opti_obj} is the sum of the terms listed under it in Table~\ref{tab:opti_costs}.

The \emph{tracking} group $E^{\mathrm T}$ anchors the solution to the IK reference, in configuration space through the joint angles and the base position, and in the world through the head and both ankles.
Those two ends of the kinematic chain are what carry posture, and the base-relative torso term is the only constraint on base orientation: without it the pelvis drifts while every other cost stays satisfied.

The \emph{contact} group $E^{\mathrm C}$ preserves the interaction.
$E^{\mathrm c}$ drives each active contact link onto its object-frame target \eqref{eq:contactpoint}, so the robot establishes contact at the same location on the object surface as the human,  independently of where the object is carried and of how it was rescaled.
A position residual leaves the link free to rotate about that location, so $E^{\mathrm{ho}}$ additionally constrains the orientation of each active palm.
The two stance terms act on the feet: $E^{\mathrm{fz}}$ and $E^{\mathrm{fo}}$ hold a supporting foot at its reference height and sole orientation, removing the float and tilt left by the per-frame IK reference.
Smoothness is the single jerk penalty $E^{\mathrm j}$.
A small weight removes the jitter of the IK reference without competing with the contact terms, while larger weights begin to erase detail of the source motion.

\begin{table}[t]
\centering
\caption{Cost terms of \eqref{eq:opti}, grouped as in \eqref{eq:opti_obj}.}
\label{tab:opti_costs}
\footnotesize
\renewcommand{\arraystretch}{1.35}
\setlength{\belowrulesep}{0pt}
\setlength{\tabcolsep}{4pt}
\begin{tabular}{l l c r}
\toprule
Cost term & Symbol & Definition & $w$ \\
\midrule
\rowcolor{black!8}
\textbf{Tracking}  & \boldmath$E^{\mathrm T}$   & \boldmath$E^\theta+E^{\mathrm b}+E^{\mathrm h}+E^{\mathrm f}+E^{\mathrm r}$ & \\
Joint position     & $E^\theta$        & $\sqnorm{\vc\theta-\vc\theta^*}$ & $1$ \\
Base position      & $E^{\mathrm b}$   & $\sqnorm{\vc p^w_b-\vc p^{w*}_b}$ & $1$ \\
Head position      & $E^{\mathrm h}$   & $\sqnorm{\vc p^w_{\mathrm{head}}-\vc p^{w*}_{\mathrm{head}}}$ & $1$ \\
Feet position      & $E^{\mathrm f}$   & $\sum_{i\in\mathcal F}\sqnorm{\vc p^w_{c,i}-\vc p^{w*}_{c,i}}$ & $100$ \\
Torso orientation  & $E^{\mathrm r}$   & $\sqfro{\Rmat^b_{\mathrm{torso}}-\Rmat^{b*}_{\mathrm{torso}}}$ & $1$ \\
\rowcolor{black!8}
\textbf{Contact}   & \boldmath$E^{\mathrm C}$   & \boldmath$E^{\mathrm c}+E^{\mathrm{ho}}+E^{\mathrm{fz}}+E^{\mathrm{fo}}$ & \\
Object contact     & $E^{\mathrm c}$   & $\sum_{i\in\mathcal C}\beta_i\sqnorm{\vc p^o_{c,i}-\vc p^{o*}_{c,i}}$ & $1000$ \\
Hand orientation   & $E^{\mathrm{ho}}$ & $\sum_{i\in\mathcal H}\beta_i\sqfro{\Rmat^w_{c,i}-\Rmat^{w*}_{c,i}}$ & $5$ \\
Stance foot height & $E^{\mathrm{fz}}$ & $\sum_{i\in\mathcal F}\gamma_i\,(z_{c,i}-z^*_{c,i})^2$ & $2000$ \\
Stance foot orient.& $E^{\mathrm{fo}}$ & $\sum_{i\in\mathcal F}\gamma_i\sqfro{\Rmat^w_{c,i}-\Rmat^{w*}_{c,i}}$ & $10$ \\
\rowcolor{black!8}
\textbf{Smoothness} & \boldmath$E^{\mathrm S}$  & \boldmath$E^{\mathrm j}$ & \\
Joint jerk         & $E^{\mathrm j}$   & $\sqnorm{\delta\ddot{\vc q}}$ & $10^{-3}$ \\
\bottomrule
\end{tabular}
\end{table}

\subsection{Contact Refinement}
\label{subsec:contactrefine}

The formulation above assumes the source annotations are accurate.
Because \eqref{eq:contactpoint} is evaluated independently at every frame, a noisy reconstruction can yield a target that shifts across the object surface from one frame to the next, so a stationary grasp is represented as a moving contact point and the arm is driven to follow it.
Under the assumption that a contact does not slip on the object, we therefore replace the per-frame targets of each contiguous contact segment by their mean, reducing the segment to a single static contact point.
This representation is also compact enough to be edited by hand: a web interface built on Viser~\cite{yi2025viser} exposes one point per segment, so contacts that penetrate the object or float above its surface can be repositioned and the optimization re-run without modifying the source data.
Such correction matters most when the source itself is uncertain, particularly for interactions reconstructed from monocular video, where occlusion and depth ambiguity frequently displace the estimated contacts (Sec.~\ref{subsec:rgbresult}).

\section{Results}
\label{sec:results}

We evaluate HOI-Retarget across different human-motion datasets, object categories,
humanoid embodiments, and downstream physics-refinement methods. 
The quantitative kinematic benchmark uses 4,421 clips from OMOMO
\cite{li2023omomo} covering 13 object categories, retargeted to the Unitree G1 \cite{unitree_g1}. 
To assess broader applicability, we retarget additional datasets \cite{kim2024parahome, zuo2026corolehoi, lu2025humoto, zhao2024imhd, zhang2023neuraldome} to both the Unitree G1 and H2 \cite{unitree_g1, unitree_h2}. 
Dynamic refinement is evaluated on the Unitree G1 using an RL tracker in Isaac Lab
\cite{mittal2025isaaclab} and the MuJoCo-based SBTO implementation of
DynaRetarget \cite{todorov2012mujoco,dhedin2026dynaretarget}.

Through these experiments, we address the following questions:
\begin{enumerate}
    \item \textit{HOI preservation:} How accurately does HOI-Retarget preserve
    the source object interaction relative to existing kinematic retargeters?
    \item \textit{Data augmentation:} Do object-frame contacts remain
    consistent as object size changes?
    \item \textit{Video retargeting:} Can the pipeline process contacts
    estimated from monocular video?
    \item \textit{Dynamic refinement:} Does the relative advantage of the
    kinematic references persist after physics-based refinement?
\end{enumerate}
All metrics are defined in Appendix~\ref{app:notation}.

Figure~\ref{fig:diverse} illustrates the breadth of the resulting kinematic
references across the six HOI sources, two humanoid embodiments, and objects
with substantially different geometry and scale. The same formulation also
processes two-person demonstrations by retargeting each actor against the
shared object trajectory, producing synchronized multi-agent references. This
establishes input and embodiment coverage at the kinematic level; mutual
collision avoidance and dynamically coupled multi-robot execution are outside
the scope of this experiment.

\begin{figure}[t]
    \centering
    \includegraphics[width=1\linewidth]{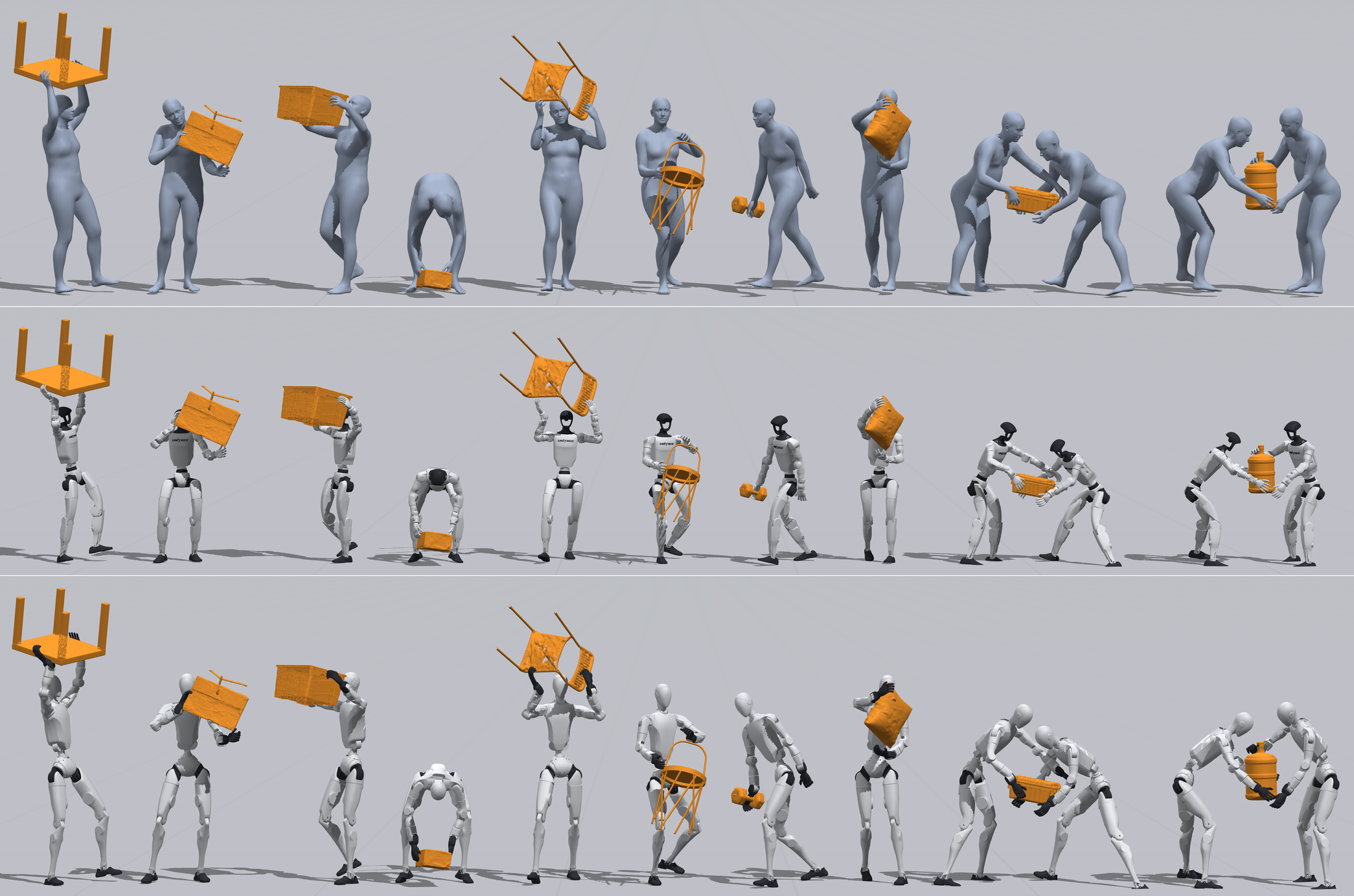}
    \caption{Representative retargets across HOI sources and embodiments. Rows
    show the source SMPL-X motion, Unitree G1, and Unitree H2.}
    \label{fig:diverse}
\end{figure}

\subsection{HOI Preservation}
\label{subsec:benchmark}

\begin{table}[b]
\centering
\caption{Kinematic retargeting on the common OMOMO subset\\ (3,997 clips; 3,604
for contact metrics).}
\label{tab:benchmark}
\footnotesize
\setlength{\tabcolsep}{4pt}
\begin{tabular}{l c @{\hspace{8pt}}|@{\hspace{8pt}} cc}
\toprule
Metric & GMR & OmniRetarget & Ours \\
\midrule
Body pose deviation ($^\circ$) $\downarrow$  & \textbf{20.0}   & 25.0   & 21.0 \\
Link vel.\ direction ($^\circ$) $\downarrow$ & \textbf{19.4}   & 28.9   & 25.2 \\
Contact-point gap (m) $\downarrow$           & 0.368  & 0.183  & \textbf{0.005} \\
Rel. hand-orient. change ($^\circ$) $\downarrow$ & \textbf{2.3} & 37.6 & 6.1 \\
Body jerk (m/s$^3$) $\downarrow$             & 66.7   & 193.5  & \textbf{49.9} \\
\midrule
Compute time (s/clip) $\downarrow$            & \textbf{9.1}    & 156.2  & 33.7 \\
\bottomrule
\end{tabular}
\end{table}

We compare HOI-Retarget with OmniRetarget
\cite{yang2025omniretarget}, the closest interaction-aware kinematic method,
and use GMR \cite{araujo2025gmr} as an object-agnostic control. GMR tracks the
human body while leaving the object interaction unconstrained and therefore
indicates the body-motion fidelity attainable when contact preservation is not
optimized. Because neither baseline resizes the object mesh, we disable our
mesh scaling for this comparison. All methods consequently manipulate the same
object geometry, while the object trajectory is adjusted only to bring it into
the robot's workspace, as done by OmniRetarget.

OmniRetarget fails to return a feasible solution for 424 of the 4,421 clips
($9.6\%$), predominantly for large, extended objects. Table~\ref{tab:benchmark}
therefore reports means over the 3,997 clips solved by all three methods, so
each method is evaluated on the same motions; the two contact metrics use the
3,604 clips in this subset that contain hand--object contact. The contact-point
gap directly evaluates the central objective of HOI-Retarget, namely recovery
of the labeled source contact location. Body-pose deviation, link-velocity
direction, and jerk quantify the corresponding effect on the transferred body
motion.

\begin{figure}[t]
    \centering
    \includegraphics[width=1\linewidth]{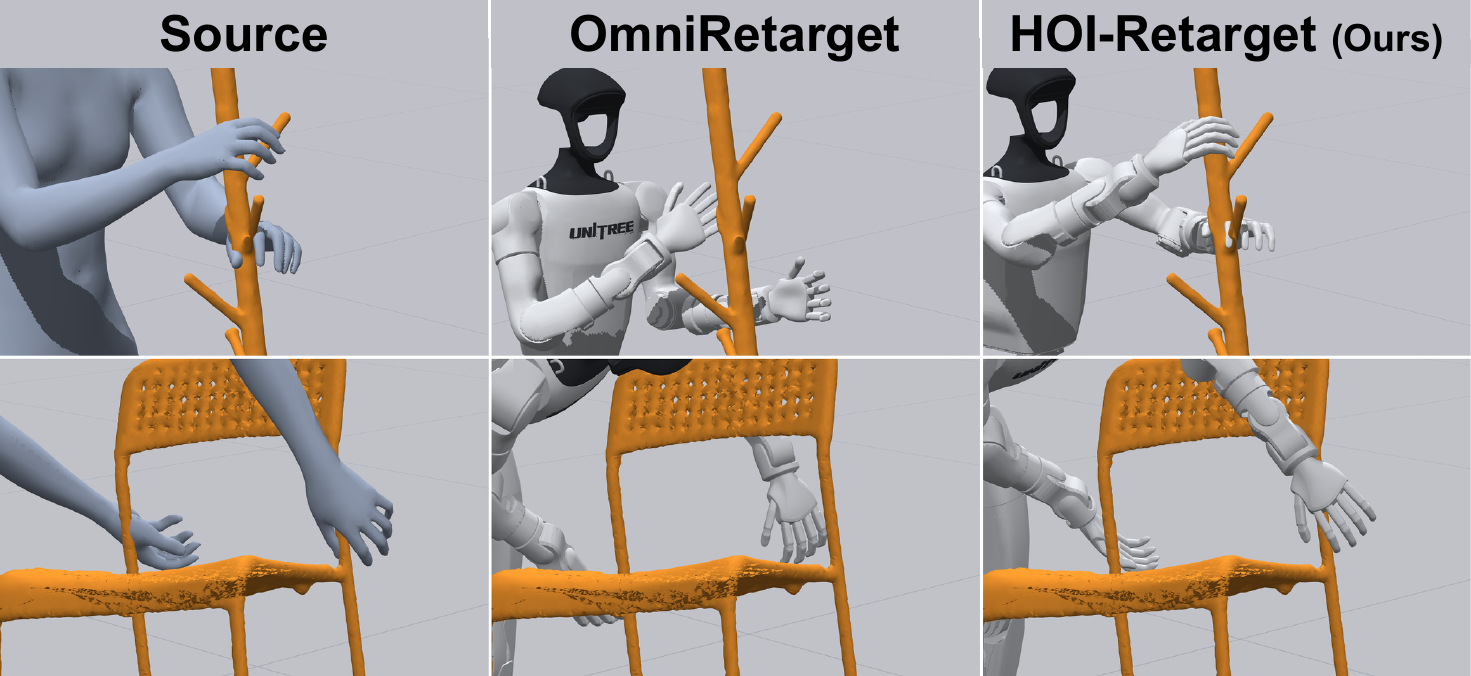}
    \caption{Qualitative comparison with OmniRetarget. HOI-Retarget recovers
    the labeled source contact regions on the coat rack and chair.}
    \label{fig:hoivsomni}
\end{figure}

HOI-Retarget reduces the mean contact-point gap from $18.3$~cm for
OmniRetarget to $0.5$~cm and the relative hand-orientation-change error from
$37.6^\circ$ to $6.1^\circ$. At the same time, it lowers body-pose deviation
from $25.0^\circ$ to $21.0^\circ$ and body jerk from $193.5$ to
$49.9$~m/s$^3$, while requiring $33.7$ rather than $156.2$ seconds per clip.
GMR attains the lowest body-pose and link-velocity-direction errors, as
expected for the object-agnostic control, but leaves a $36.8$~cm contact gap.
Its low orientation-change error must likewise be interpreted together with
this gap: it reproduces the hand's rotational evolution without placing the
hand at the intended object contact. The examples in
Fig.~\ref{fig:hoivsomni} visualize this distinction: OmniRetarget preserves the
broad body--object arrangement, whereas HOI-Retarget places the palms at the
labeled source regions on the object.

\subsection{Data Augmentation}
\label{subsec:dataaug}

\begin{figure}[t]
    \centering
    \includegraphics[width=1\linewidth]{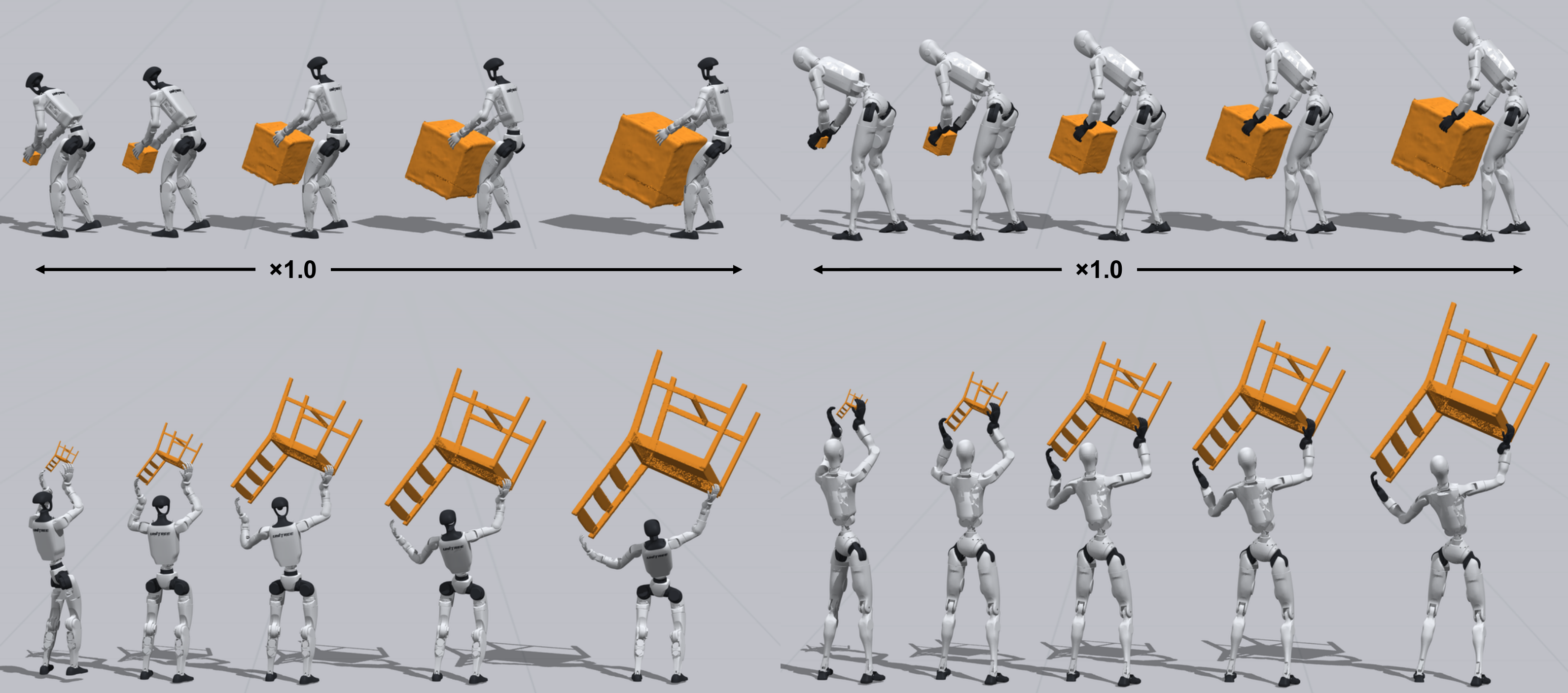}
    \caption{Contact-preserving object-scale augmentation over the tested
    range $\times0.25$--$\times1.50$ on the G1 and H2.}
    \label{fig:aug}
\end{figure}

We evaluate object-scale augmentation by varying the additional scale
multiplier from $\times0.25$ to $\times1.50$ and re-solving the same source
interaction without changing its contact annotation. As shown for box and
chair interactions on the G1 and H2 in Fig.~\ref{fig:aug}, the palms follow
the corresponding surface regions as the object becomes smaller or larger,
rather than remaining at their original world positions. This qualitative
result demonstrates geometric contact consistency over the tested scale range;
it does not by itself establish dynamic feasibility for every generated size.

\subsection{Retargeting from Video}
\label{subsec:rgbresult}

\begin{figure}[t]
    \centering
    \includegraphics[width=1\linewidth]{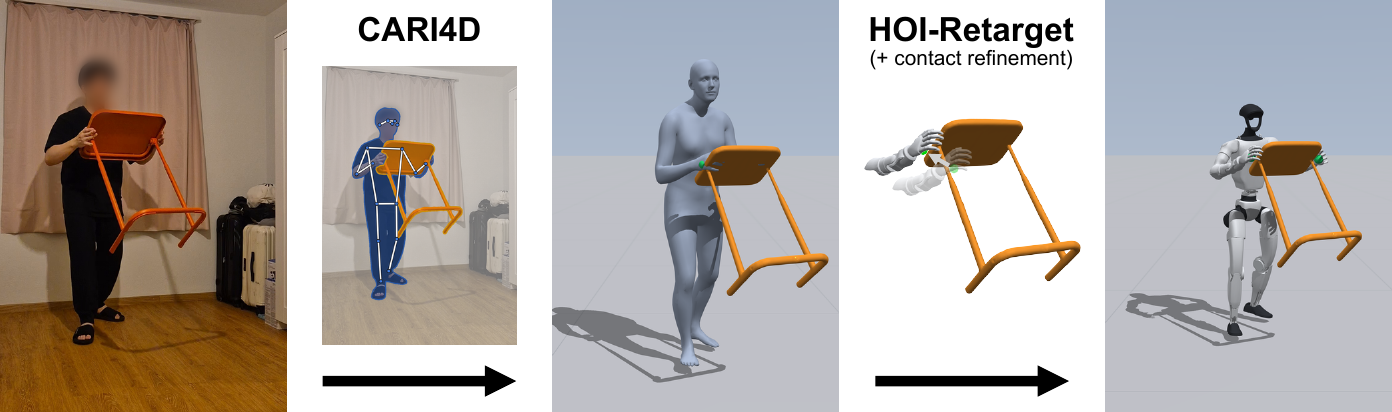}
    \caption{Example of using HOI-Retarget from video reconstruction using CARI4D. Contact refinement tool has been used to account for monocular reconstruction artifacts.}
    \label{fig:videorecon}
\end{figure}

We further test whether the pipeline can consume an HOI source reconstructed outside a motion-capture setup (Fig.~\ref{fig:videorecon}). 
A subject is recorded with a monocular camera while moving a table not contained in the datasets above, and CARI4D \cite{cari4d2026} reconstructs the human mesh, object mesh, and their trajectories. 
In this example, occlusion and depth ambiguity cause the estimated hand contact to alternate between penetration and separation, which would carry over after retargeting. 
We hence collapse each sticking-contact segment to one point and manually correct the remaining displaced target before applying the same retargeting formulation. 
This experiment qualitatively establishes compatibility with a monocular reconstruction pipeline; it does not evaluate reconstruction accuracy across a video dataset.

\subsection{Dynamic Refinement}
\label{subsec:dynamicrefinement}

\begin{table}[b]
\centering
\caption{Dynamic-refinement quality on the common subset\\ (68 clips; 63 for
contact metrics).}
\label{tab:dynrefablation}
\footnotesize
\setlength{\tabcolsep}{6pt}
\begin{tabular}{@{}l r r r r@{}}
\toprule
& \multicolumn{2}{c}{RL Tracker} & \multicolumn{2}{c}{SBTO} \\
\cmidrule(lr){2-3}\cmidrule(lr){4-5}
Metric & Omni. & \textbf{Ours} & Omni. & \textbf{Ours} \\
\midrule
Body pose deviation ($^\circ$) $\downarrow$         & 26.5  & \textbf{23.5}  & 26.0  & \textbf{23.1} \\
Link vel.\ direction ($^\circ$) $\downarrow$        & 47.9  & \textbf{46.7}  & 51.2  & $50.8^{\ast}$ \\
Object path err.\ (m) $\downarrow$                  & 0.285 & \textbf{0.213} & 0.269 & \textbf{0.194} \\
Contact-point gap (m) $\downarrow$                  & 0.096 & \textbf{0.086} & 0.141 & \textbf{0.099} \\
Rel. hand-orient. change ($^\circ$) $\downarrow$    & 31.5  & \textbf{25.3}  & 38.4  & \textbf{28.1} \\
Body jerk (m/s$^3$) $\downarrow$                    & 90.8  & \textbf{85.8}  & 58.2  & $55.4^{\ast}$ \\
\bottomrule
\end{tabular}
\par\vspace{1pt}\scriptsize $^{\ast}$Difference not significant under a paired Wilcoxon test.
\end{table}

Finally, we test whether the difference between the kinematic references
remains observable after enforcing simulated dynamics. We apply two distinct
refiners to references produced by OmniRetarget and HOI-Retarget: a per-clip
policy trained with PPO \cite{schulman2017ppo} through RSL-RL
\cite{schwarke2025rslrl,rudin2022massively} in Isaac Lab \cite{mittal2025isaaclab}, and the
MuJoCo-based SBTO method of DynaRetarget
\cite{todorov2012mujoco,dhedin2026dynaretarget}. The RL tracker uses
DeepMimic-style body- and object-tracking rewards \cite{peng2018deepmimic,weng2025hdmi},
privileged observations, and no domain randomization. It is therefore used to
generate physically consistent simulation trajectories, not as a deployable
controller. For each refiner, both reference types receive the same training
or optimization budget.

Table~\ref{tab:dynrefablation} reports the common subset of 68 clips for which
both reference types completed both refinement procedures; the two contact
metrics use the 63 clips that contain hand--object contact. Under RL
refinement, our references reduce object-path error from $0.285$ to $0.213$~m,
contact-point gap from $9.6$ to $8.6$~cm, and relative hand-orientation-change
error from $31.5^\circ$ to $25.3^\circ$. Under SBTO, the corresponding values
decrease from $0.269$ to $0.194$~m, $14.1$ to $9.9$~cm, and $38.4^\circ$ to
$28.1^\circ$. The SBTO differences in link-velocity direction and jerk are not
statistically significant under the paired Wilcoxon test and are therefore not
interpreted as improvements. Although dynamic refinement necessarily moves the
motion away from its kinematic reference, both refiners retain the relative
advantage in object path, contact location, and hand orientation. These results
support the use of HOI-Retarget as a more accurate initialization for physics
grounding on the common successfully refined subset.

\section{Discussion}
\label{sec:discussion}

We presented HOI-Retarget, a contact-centric method for transferring human--object interactions to humanoid robots. Object-frame contact targets and windowed trajectory optimization preserve where and when the human touches the object while accommodating differences in embodiment and object size. On 13 OMOMO object categories, HOI-Retarget reduces the mean contact-point gap of the closest interaction-aware kinematic baseline from $18.3$ to $0.5$~cm at less than one quarter of its compute, and the relative advantage remains after both RL- and optimization-based dynamic refinement. The same formulation supports object-scale augmentation, monocular-video input, and synchronized multi-agent references. We release the pipeline and 6,952 kinematic humanoid interaction references spanning 75 unique objects.

\subsection{Limitations}

\textit{Reference quality.}
\label{subsec:referencequality}
HOI-Retarget preserves the source motion and contacts without reasoning about the intent of the interaction, so errors in the input can propagate to the result. 
Motion capture and monocular reconstruction may contain unnatural poses, jitter, inaccurate contacts, or body--object penetration.
Contact-segment refinement addresses erroneous contact annotations, but it does not correct the underlying human or object motion.

\textit{Dexterity.}\label{subsec:nohand}
The current representation models each hand as a palm contact and therefore captures where and when the hand meets the object, but not finger closure, contact distribution, or grasp forces. Consequently, interactions that can be supported between the palms or against the body are suitable for dynamic refinement, whereas tasks requiring articulated grasping remain kinematic references.

\subsection{Future Work}
\label{subsec:futureworks}

A primary extension is grasp-aware retargeting for whole-body dexterous interaction. Rather than constraining only palm locations, the trajectory optimization could be conditioned on specific grasp configurations that couple finger posture, object-relative contacts, and whole-body motion. This would allow the pipeline to preserve the demonstrated interaction while adapting the grasp to the geometry and kinematics of an articulated robotic hand. A secondary extension is to incorporate the body--object collision cost of Appendix~\ref{app:horizon} as an optional non-penetration objective. Because the generated motions are intended as references for RL or sampling-based dynamic refinement, eliminating every small penetration at the kinematic stage is not essential: simulation-based refinement can resolve many such inconsistencies while enforcing dynamics. Collision-aware retargeting is nevertheless valuable when severe source penetration would otherwise produce an unsuitable initialization.

\clearpage 

\appendices

\section{Notation and Evaluation Metrics}
\label{app:notation}

Table~\ref{tab:optA_symbols} lists the symbols used throughout and Table~\ref{tab:metrics} defines the reported metrics.

\begin{table}[h]
\centering
\caption{Symbols used for HOI-Retarget.}
\label{tab:optA_symbols}
\footnotesize
\renewcommand{\arraystretch}{1.15}
\begin{tabular}{@{}l l@{}}
\toprule
Symbol & Description \\
\midrule
$(\cdot)^*$               & Reference quantity \\
$(\cdot)^{\mathrm h}$     & Human source quantity \\
$(\cdot)^w,(\cdot)^b,(\cdot)^o$ & World, base, object frame \\
$\delta(\cdot)$           & Backward difference, \\
                          & \quad $\delta(\cdot)_t=(\cdot)_t-(\cdot)_{t-1}$ \\
$\vc q$                   & Configuration $(\vc p^w_b,\Rmat^w_b,\vc\theta)$ \\
$\ddot{\vc q}$            & Acceleration, $\delta^2\vc q/\Delta t^2$ \\
$\vc p_b,\Rmat_b$         & Base position, orientation \\
$\vc\theta$               & Joint angles \\
$\vc\theta_{\min},\vc\theta_{\max}$ & Joint position limits \\
$\dot{\vc\theta}_{\mathrm{lim}}$ & Joint velocity limits \\
$\vc p_o,\Rmat_o$         & Object pose (fixed input) \\
$\vc p_{c,i},\Rmat_{c,i}$ & Pose of contact link $i$ \\
$z_{c,i}$                 & Height of contact link $i$ \\
$\beta_i,\gamma_i$        & Contact, stance flag of link $i$ \\
$\mathcal C,\mathcal H,\mathcal F$ & Contact links, palms, feet \\
$\mathcal W_k$            & $k$-th solve window, $[\,t_k,t_k{+}H\,)$ \\
$H,p$                     & Window length, pinned frames \\
$\Delta t$                & Timestep \\
$\norm{\cdot}_F$          & Frobenius norm \\
\bottomrule
\end{tabular}
\end{table}

\begin{table}[h]
\centering
\caption{Evaluation metrics reported in Tables~\ref{tab:benchmark} and~\ref{tab:dynrefablation}.}
\label{tab:metrics}
\footnotesize
\renewcommand{\arraystretch}{1.25}
\begin{tabular}{@{}>{\raggedright\arraybackslash}p{0.30\columnwidth} >{\raggedright\arraybackslash}p{0.64\columnwidth}@{}}
\toprule
Metric & Definition \\
\midrule
Body pose deviation ($^\circ$) &
  Mean over frames of the mean angle between the robot's and the human's $16$ bone directions, taken in the pelvis frame for the legs and the trunk frame for the arms. \\

Link vel.\ direction ($^\circ$) &
  Mean angle between the robot's and the human's object-relative link velocity $\Rmat^{w\top}_o(\dot{\vc p}^w-\dot{\vc p}^w_o)$, over the palms, ankles, head and pelvis, where both exceed $2$~cm/s. \\

Contact-point gap (m) &
  Mean distance on the object between the robot's palm and the human's contact point \eqref{eq:contactpoint}, over frames in contact. This is the residual $E^{\mathrm c}$ minimizes. \\

Rel.\ hand-orient.\ change ($^\circ$) &
  Mean angle between the robot's and the human's change in palm orientation in the object frame since the first frame of the contact segment. \\

Object path err.\ (m) &
  Mean distance between the robot-side and the human-side object translation, each taken relative to its own first frame. \\

Body jerk (m/s$^3$) &
  Mean magnitude of the third difference of world link position, over links and frames. \\

Compute time (s/clip) &
  Mean wall-clock time of a complete single-threaded run per clip, on one core of AMD EPYC 7H12 node. \\
\bottomrule
\end{tabular}
\end{table}

\section{Optimization Horizon}
\label{app:horizon}

The windowed formulation of Sec.~\ref{subsec:windowedto} sits between the per-frame solves used by prior kinematic retargeters and a single program over the whole clip.
Its cost is visible only on objectives whose size grows with the interaction, so we evaluate the horizon with and without an optional body--object collision cost, which penalizes each of $17$ robot capsules against a sampled object surface and stands for the non-convex terms a bounded horizon is meant to make affordable.
We take the ten longest clips of each of the 13 OMOMO objects ($130$ clips, $44{,}974$ frames) at full object mesh scale, and give every solve one CPU and $16$~GB of memory.
Table~\ref{tab:horizon} reports solve outcomes over all $130$ clips and every other column as a median over the $87$ that all five configurations completed, so that no arm is scored on an easier subset than its row-mates.

Per-frame solves are the cheapest and by far the worst, because a single free frame cannot trade a contact against its neighbors and degrades the jerk term, here the RMS third difference in joint space, to a first-order velocity penalty.
Windowing gives up $0.9^\circ$ of link velocity direction and 66 rad/s$^3$ of joint jerk against a full-trajectory solve, at lower time and with the memory it adds on top of the IK stage smaller.
That margin decides feasibility once the collision cost is present: every windowed solve completes, while $43$ of the $130$ full-trajectory solves are killed for exceeding the memory budget.
On the $87$ clips both complete, the full-trajectory horizon costs $2.2\times$ the memory for $1.3\times$ the time and buys little in return, matching windowing to $0.1^\circ$.
Given $64$~GB all $43$ solve, peaking at $45.9$~GB against windowing's $14.4$~GB, so the claim is feasibility under a fixed allocation rather than impossibility; and only the second stage is bounded in clip length, since the IK of Sec.~\ref{subsec:ikretargetobjscale} still grows linearly with it.

\begin{table}[h]
\centering
\caption{Optimization horizon, with and without the collision cost.}
\label{tab:horizon}
\footnotesize
\setlength{\tabcolsep}{4pt}
\begin{tabular}{@{}l c c r r c c@{}}
\toprule
Horizon & Coll. & Solved & \shortstack{Time\\ (s)} & \shortstack{RAM\\ (MB)} & \shortstack{Link vel.\ dir.\\ ($^\circ$)} & \shortstack{Joint jerk\\ (rad/s$^3$)} \\
\midrule
Per-frame & \xmark & 130/130 & 37.7 & 1440 & 40.3 & 588 \\
\midrule
\multirow{2}{*}{Windowed} & \xmark & 130/130 & 39.4  & 1763 & 24.4 & 251 \\
                          & \cmark & 130/130 & 159.9 & 3876 & 24.5 & 257 \\
\midrule
\multirow{2}{*}{Full traj.} & \xmark & 130/130 & 45.8  & 2400 & 23.5 & 185 \\
                            & \cmark & 87/130  & 210.5 & 8671 & 24.4 & 186 \\
\bottomrule
\end{tabular}
\end{table}

\section*{Acknowledgments}
The authors used generative AI tools to assist with language refinement and code generation.

\clearpage

\bibliographystyle{IEEEtran}
\bibliography{references}

\end{document}